\documentclass[letterpaper, 10 pt, conference]{ieeeconf}  %

\IEEEoverridecommandlockouts                              %

\usepackage{graphics} %
\usepackage{epsfig} %
\usepackage{times} %
\usepackage{amsmath} %
\usepackage{amssymb}  %
\usepackage{cite}
\usepackage[font=footnotesize]{caption}
\usepackage{pifont}%
\usepackage{multirow}
\usepackage{hyperref}
\hypersetup{
    colorlinks=true,
    linkcolor=blue,
    filecolor=magenta,      
    urlcolor=cyan,
    pdftitle={Overleaf Example},
    pdfpagemode=FullScreen,
    }
\usepackage{bm}

\usepackage[dvipsnames,table]{xcolor}

\title{\Large \bf
OceanSim: A GPU-Accelerated Underwater Robot Perception Simulation Framework
}

 \author{%
Jingyu Song$^{*1}$, Haoyu Ma$^{*1}$, Onur Bagoren$^{1}$, Advaith Sethuraman$^{1}$, Yiting Zhang$^{1}$, and Katherine A. Skinner$^{1}$
    \thanks{$^*$denotes equal contribution}
    \thanks{$^1$J. Song, H. Ma, O. Bagoren, A. Sethuraman, Y. Zhang and K. A. Skinner are with the Department of Robotics, University of Michigan, Ann Arbor, MI 48109 USA. Corresponding author e-mail: \tt\small jingyuso@umich.edu}
 }

\date{March 2025}

\begin{document}

\maketitle

\begin{abstract}
Underwater simulators offer support for building robust underwater perception solutions. Significant work has recently been done to develop new simulators and to advance the performance of existing underwater simulators. Still, there remains room for improvement on physics-based underwater sensor modeling and rendering efficiency. In this paper, we propose OceanSim, a high-fidelity GPU-accelerated underwater simulator to address this research gap. We propose advanced physics-based rendering techniques to reduce the sim-to-real gap for underwater image simulation. We develop OceanSim to fully leverage the computing advantages of GPUs and achieve real-time imaging sonar rendering and fast synthetic data generation. We evaluate the capabilities and realism of OceanSim using real-world data to provide qualitative and quantitative results. %
The code and detailed documentation are made available on the project website to support the marine robotics community: \href{https://umfieldrobotics.github.io/OceanSim}{https://umfieldrobotics.github.io/OceanSim}.

\end{abstract}

\section{Introduction}
\label{Sec:introduction}

Marine robotic platforms support a wide range of applications, including marine exploration, underwater infrastructure inspection, and ocean environment monitoring~\cite{rahman2019svin2, zhang2024recgs, zhao2023tightly, kim2013real, bagoren2025pugs}. Advancing autonomous capabilities for marine robotics offers potential to reduce time and cost of carrying out these critical missions.  %
To achieve this, a fundamental requirement is reliable underwater robot perception, which allows robots to sense and understand their surrounding environments, detect nearby objects, and navigate complex subsea terrains autonomously.

Still, underwater robot perception faces unique challenges compared to perception systems operating in terrestrial environments. %
Conducting underwater robot surveys to collect real-world data is usually resource-intensive, requiring sophisticated hardware, controlled experimental setups, and extensive field work. Furthermore, real-world testing in underwater environments can be impractical, hazardous, or expensive. To address these challenges, the development of underwater simulation platforms has emerged as a promising approach~\cite{Holoocean, potokar2022holoocean, UNav_sim, UWsim}. Underwater simulators can provide great value to researchers and engineers by offering a controlled, repeatable, and scalable environment to develop, test, and refine perception algorithms before deploying them in real-world scenarios.

\begin{figure}[t]
    \centering
    \includegraphics[width=1.0\linewidth]{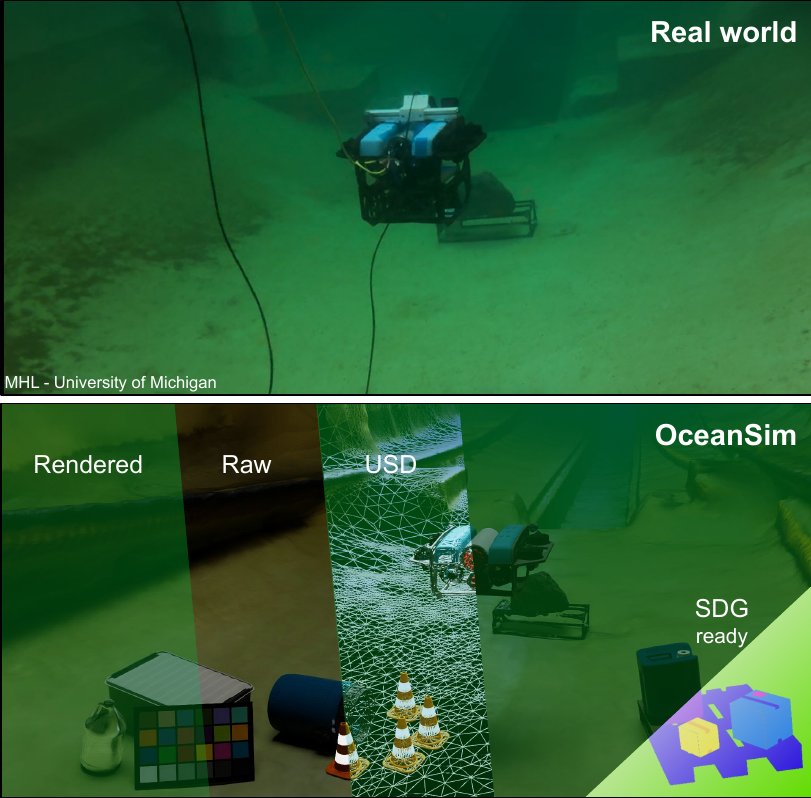}
    \caption{We propose OceanSim to improve the fidelity and speed of underwater rendering techniques by leveraging GPU acceleration. OceanSim is ready for large-scale synthetic data generation (SDG) and is able to reconstruct realistic digital twins to mimic real-world robot missions with great flexibility to customize the scene.}
    \label{fig:pitch}
    \vspace{-6mm}
\end{figure}

Motivated by this pressing need, researchers in~\cite{UWsim} proposed one of the first open-source underwater simulation tools called UWSim, which supports testing and integrating perception and control algorithms on simulated marine robot platforms and provides a user-friendly interface developed using the Robot Operating System (ROS)~\cite{ros}. UWSim~\cite{UWsim} was followed by several works~\cite{UUV_sim, UW_Morse, URsim, Stonefish, sim2vita} that contributed additional simulation capabilities to the community. Despite these advances, current underwater robotics simulators are limited in that they lack GPU-accelerated rendering~\cite{UWsim, UUV_sim}, provide low fidelity in rendering vision sensors~\cite{UW_Morse, URsim}, or require sub-optimal dependencies on non-scalable libraries~\cite{Stonefish, sim2vita} to simulate sensors.
More recently, HoloOcean~\cite{Holoocean} and UNav-Sim~\cite{UNav_sim} propose novel uses of Unreal Engine~\cite{unrealengine} to improve the fidelity of underwater simulators and to enable GPU acceleration for image rendering. However, HoloOcean~\cite{Holoocean} does not accurately model water column effects in image rendering and uses an octree-based pipeline to render sonar sensors, leaving room for achieving more realistic underwater image rendering and optimizing sonar rendering efficiency. Moreover, UNavSim~\cite{UNav_sim} lacks support for simulating acoustic sensors such as imaging sonars and Doppler Velocity Logs (DVLs), which are important for underwater scene understanding and navigation.

Therefore, there is still a need for next-generation underwater simulators that offer high-fidelity sensor rendering, efficient sonar data generation accelerated by GPUs, and advanced scalability and versatility for customizing the scenes to mimic real-world missions. Motivated by this, we propose OceanSim (Fig.~\ref{fig:pitch}), a high-performance underwater simulator accelerated by NVIDIA parallel computing technology. We build OceanSim upon NVIDIA Isaac Sim~\cite{issacsim} to leverage its high-fidelity physics-based rendering and GPU-accelerated real-time ray tracing. Notably, OceanSim bridges the field of underwater simulation and the fast-growing NVIDIA Omniverse ecosystem~\cite{Nvidia_omniverse}, facilitating the application of the vast amount of existing sim-ready assets and robot learning works in the underwater robotics field~\cite{gaussianSplatting_Isaac, object_detection_Isaac,DRL_Isaac,RL_Isaac,humanoid_RL_Isaac}.

To summarize, the main contributions of OceanSim are as follows:

\begin{itemize}
    \item OceanSim incorporates advanced physics-based rendering techniques to accurately model underwater visual and acoustic sensors.
    \item OceanSim achieves significantly faster sonar rendering speeds compared to alternative simulation engines.
    \item OceanSim maintains a versatile design as an extension to NVIDIA Issac Sim~\cite{issacsim}. By embracing the fast-growing NVIDIA Omniverse ecosystem~\cite{Nvidia_omniverse}, our simulator naturally bridges the gap between simulation for robot learning and underwater robotics.
    \item OceanSim will be available open-source to allow the research community to contribute in a collaborative paradigm.
\end{itemize}

\begin{figure*}[t]
    \centering
    \includegraphics[width=1.0\linewidth]{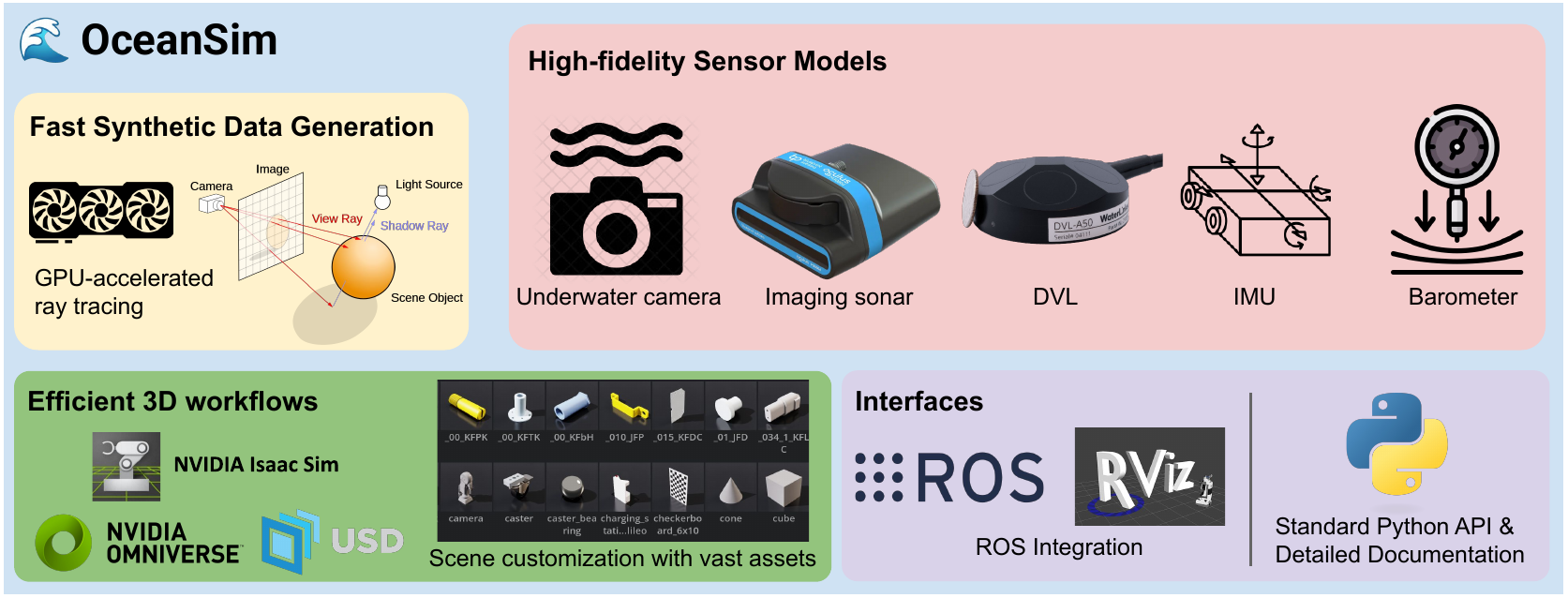}
    \caption{An overview of OceanSim, a high-performance simulator built for GPU-accelerated synthetic data generation with proposed high-fidelity underwater sensor models. OceanSim is compatible with popular 3D workflows that offer a high level of flexibility in creating digital twins, and provides user-friendly APIs and documentation.}
    \label{fig:ovceansim_overview}
\end{figure*}

We evaluate OceanSim qualitatively and quantitatively and provide comparison to real-world data. Code and detailed documentation will be released on the project page: \href{https://umfieldrobotics.github.io/OceanSim}{https://umfieldrobotics.github.io/OceanSim}.

\section{Related Works}

\subsection{Underwater Simulators}
\label{sec:related-works-underwater-simulator}
One of the growing trends in underwater robotics is to develop high-performance underwater simulators, since real-world testing in underwater environments can be challenging and resource-intensive. Early-stage underwater simulators~\cite{UWsim, UUV_sim, URsim, PBS_simulation, fluid_Gazebo} are built upon ROS~\cite{ros}. Despite the benefits of the fluid modules in ROS Gazebo that can simulate hydrodynamics, ROS is not optimized for high-fidelity visual rendering, leaving a considerably large sim-to-real gap in rendered camera images.

In recent years, game engine-powered simulators (HoloOcean\cite{Holoocean}, UNav-Sim\cite{UNav_sim}) have achieved improved rendering quality by leveraging Unreal Engine~\cite{unrealengine}. Still, the official release version of HoloOcean~\cite{Holoocean} relies on Unreal Engine 4, which does not support rendering depth images. This limits the possibility of applying physics-based models for realistic underwater image rendering. HoloOcean also lacks GPU-based sonar rendering, which leads to slow rendering time for synthetic sonar data generation. Furthermore, HoloOcean requires building an Octree cache for sonar rendering, which takes extended time when a new environment is loaded. UNav-Sim~\cite{UNav_sim} employs a newer version of Unreal Engine~\cite{unrealengine}, UE5.1, to achieve more realistic underwater image rendering. However, the image formation model is difficult to customize and has not been validated against real underwater imagery. In addition, there is no example underwater scene configuration provided in UNav-Sim, making it more difficult for users to construct an underwater digital twin. Furthermore, UNav-Sim does not support simulating acoustic sensors such as imaging sonars, which are important for underwater perception. Therefore, there still remains a gap in development of a high-performance underwater robot perception simulator that can efficiently generate high-fidelity multi-modal data.

\subsection{NVIDIA Isaac Sim}
NVIDIA Isaac Sim~\cite{issacsim} is a high-performance simulator designed for GPU-accelerated robotics simulation with superior physics accuracy and efficient tools for generating synthetic data. NVIDIA Isaac Sim has enabled successful digital twin construction and sim-to-real transfer in a wide range of fields~\cite{Automation_production_Isaac,automation_factory_Isaac, manipulation_Isaac, manipulation_training_Isaac, grasp_Isaac, gaussianSplatting_Isaac,object_detection_Isaac,semantics_Isaac, DRL_Isaac,RL_Isaac,humanoid_RL_Isaac, zhang2025xpg}. Key features of NVIDIA Issac Sim include physics-based photorealistic rendering, seamless integration with NVIDIA Omniverse~\cite{Nvidia_omniverse} and OpenUSD~\cite{OpenUSD} ecosystems, developer-friendly 3D workflows, and GPU-based ray tracing. Recently, MarineGym has extended NVIDIA Issac Sim to the underwater domain~\cite{chu2024marinegym}. MarineGym focuses on characterizing the simulation capability of NVIDIA Issac Sim for robot controls tasks and does not provide open-source implementation. On the other hand, our proposed simulation framework, OceanSim, focuses on building a high-performance simulator for underwater robot perception, and contributes novel implementations of common underwater sensors in NVIDIA Isaac Sim. OceanSim will be released as open-source to support the underwater robotics research community.

\section{Technical Approach}
OceanSim is developed as a custom extension that can be directly loaded in NVIDIA Issac Sim~\cite{issacsim}. As shown in Fig.~\ref{fig:ovceansim_overview}, OceanSim integrates 3D environment models captured in the real world and high-fidelity sensor models for cameras, imaging sonars, barometers, and DVLs. Being built upon NVIDIA Isaac Sim, OceanSim directly inherits the advantages of real-time GPU-accelerated ray tracing, and the wide range of extensions and assets shared by the rapidly growing NVIDIA Omniverse ecosystem. In practice, OceanSim users can easily customize both the virtual underwater environment and the robot perception sensor suite. %

\subsection{Sensor Models}
As shown in Fig.~\ref{fig:ovceansim_overview}, we develop several underwater sensor models for OceanSim to complement the built-in models in NVIDIA Isaac Sim. 
Specifically, OceanSim includes an image formation model that captures water column effects, with the flexibility to model several water types \cite{akkaynak_what_2017}. 
We also propose a sonar model with GPU-based ray tracing and realistic sonar noise modeling to enable fast and realistic sonar simulation. 
In addition, we design a DVL model that can simulate range-dependent adaptive frequency and dropout behaviors, which are important for testing state estimation algorithms. 
To simulate a barometer, we compute the pressure based on user-defined atmospheric pressure and water density with optional Gaussian noise.
Lastly, we use the built-in model provided by NVIDIA Isaac Sim to simulate the IMU.
The remainder of this section discusses further details for the camera, imaging sonar, and DVL sensor models.
\subsubsection{Camera}
\label{sec:sensor-model-camera}
\begin{figure}[t]
    \centering
    \includegraphics[width=1.0\linewidth]{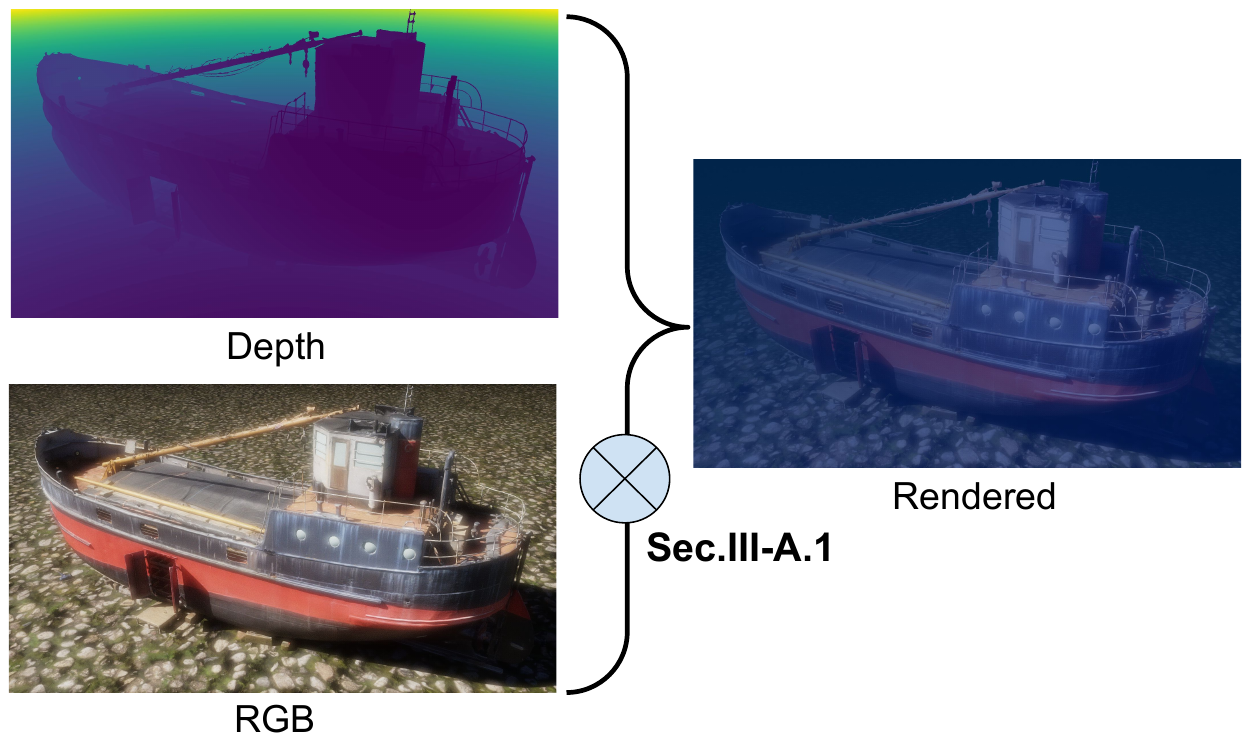}
    \caption{Image formation model integrated in OceanSim. The model takes in RGB and depth images to render underwater images. We implement the computation on the GPU.}
    \label{fig:camera_comparison}
    \vspace{-6mm}
\end{figure}

Underwater imaging is significantly affected by the water medium, primarily due to two phenomena: light attenuation and backscattering~\cite{akkaynak_revised_2018}. 
To simulate these effects, we provide a novel implementation of the underwater image formation model in NVIDIA Isaac Sim following \cite{akkaynak_revised_2018}.
We present an illustration of our underwater image rendering process in Fig.~\ref{fig:camera_comparison}. 
Specifically, we simulate water effects on the in-air image $J$ with user-defined parameters.
For each color channel \( c \in \{R, G, B\} \), the observed underwater image \( I_c \) is modeled as
\begin{equation}
\label{eq:uw_model}
I_c = J e^{-\beta_\text{attn,c}  d} + B_{\infty,c} \left(1-e^{-\beta_\text{bs,c} d} \right),
\end{equation}
where $J$ represents the in-air image rendered using the rendering engine from Isaac Sim, $d$ is the depth image, and $\beta_\text{attn,c}, \beta_\text{bs,c}$, and $B_{\infty,c}$ are the per channel attenuation, back-scatter, and veiling light components, respectively.

The model comprises two primary components, attenuation and backscattering. The attenuation term, $ J e^{-\beta_\text{attn,c}  d}$, represents the decay of the original image signal as the light propagates through water, where degradation increases with increased depth, or range, from the camera, $d$. 
The backscatter term, $B_{\infty,c} \left(1-e^{-\beta_\text{bs,c} d}\right) $, is an additive term that accounts for the light scattered by suspended particles in water, which also increases with depth from the camera and adds a veiling luminance to the image.

\begin{figure}[t]
    \centering
    \includegraphics[width=1.0\linewidth]{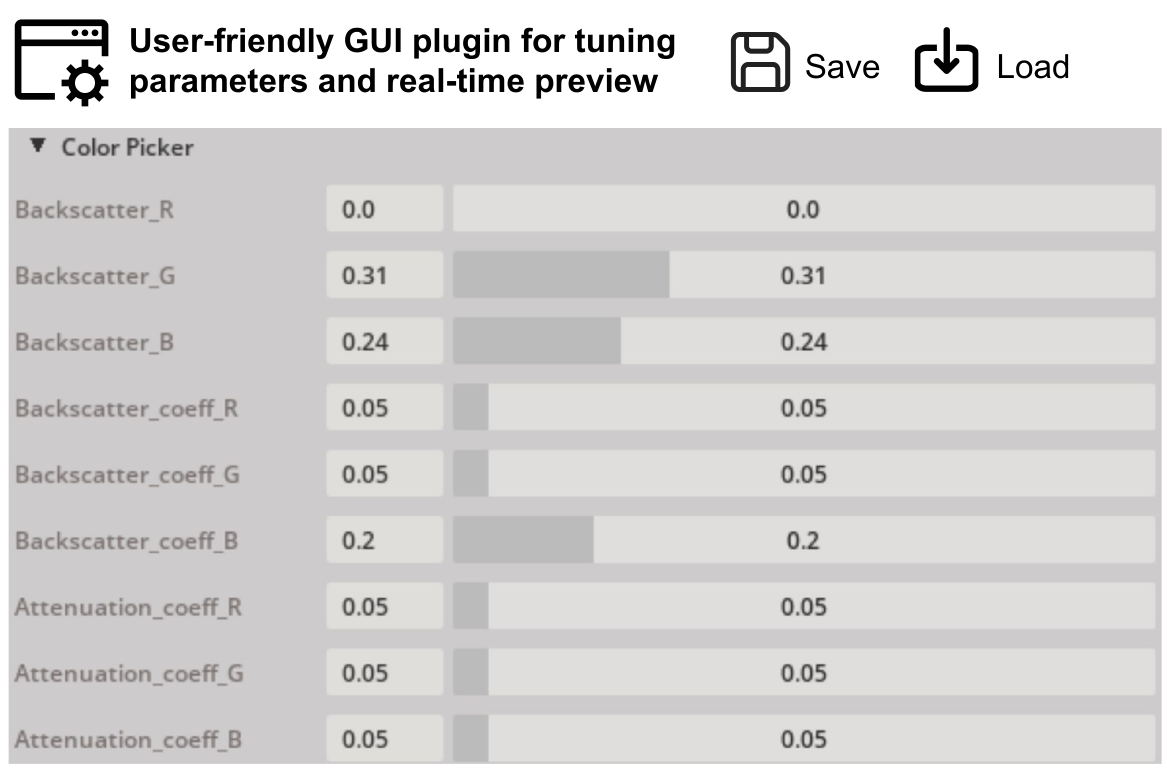}
    \caption{We provide a user-friendly GUI plugin to tune the parameters with real-time previews. Users can save the parameters to a config file that can be loaded when initializing a camera sensor.}
    \label{fig:gui_image_params}
    \vspace{-2mm}
\end{figure}

We recognize the importance of adjusting the above parameters to reflect varying water column effects that a user may want to simulate. 
Therefore, we provide a user-friendly GUI plugin in OceanSim (Fig.~\ref{fig:gui_image_params}) that allows users to configure these parameters interactively and visualize the generated images in real-time using GPU-accelerated rendering capabilities.
The tuned parameters can be saved as a configuration file that can be loaded when initializing a camera model in the simulator.

\subsubsection{Imaging Sonar}
\label{sec:sensor-model-sonar}

Sonar is an important perception sensor for underwater robots to understand the surrounding environment. Sensors like imaging sonar are able to provide time-of-flight information where optical imagery may fail, such as in high turbidity or low lighting conditions. This work explores novel GPU-accelerated solutions to solve the performance bottleneck of sonar rendering observed in prior work~\cite{potokar2022holoocean, Holoocean}.

Specifically, we leverage Omniverse Replicator~\cite{OmniverseReplicator}, a dedicated tool developed by NVIDIA for fast synthetic data generation, to develop our sonar rendering module. We set up a virtual rendering viewport and attach the Omniverse Replicator tool to it, which enables us to retrieve the scene geometry information including normals and semantics within a pre-defined field of view using GPU acceleration. Internally, scene geometry is queried using ray-tracing, similar to prior sonar simulators ~\cite{raytrace_sonar_sim}. We use a built-in API called the point cloud annotator to obtain the scene geometry. In the retrieved point cloud, the formula used to compute the intensity of each returned data point is as follows\cite{demarco2015computationally}:

\begin{equation}
    I_{sonar} = A_r \left(- \frac{\vec{v}_{in}}{|\vec{v}_{in}|} \cdot \frac{\vec{v}_{n}}{|\vec{v}_{n}|}\right)e^{-\alpha d},
\end{equation}
where $v_{in}$ denotes the incident vector to the data point and $v_n$ is the normal vector \cite{potokar2022holoocean}. The dot product regulates the received intensity based on variations of the surface normal. The exponential term applies range-dependent intensity attenuation based on attenuation coefficient $\alpha$ and distance $d$ from the sensor. Finally, $A_r$ represents the acoustic reflectance of the queried point, which depends on material properties. In OceanSim, we offer two options for defining material properties. By default, a uniform $A_r$ is set for an entire object. Alternatively, we design a pipeline leveraging the built-in Isaac Sim Semantics Schema Editor to allow users to customize the acoustic reflectivity of objects in the scene with greater flexibility by assigning material properties to the semantic model.

The rendering of OceanSim's imaging sonar follows that of HoloOcean \cite{potokar2022holoocean}. After obtaining the set of queried returns $\mathcal{Q} = \{x_i \in \mathbb{R}^3  | \ x_i \text{ in view and not occluded} \}$ where $x_i$ is a point in the scene, we bin all the returns on a range-azimuth polar grid by summing up intensity measurements that fall into each grid. Then, we add additive and multiplicative noise to the binned value to model the speckle noise properties of real imaging sonar. We begin with the additive noise term $\tilde{w}^{sa}$: 

\begin{equation}
\tilde{w}^{sa} = \frac{r_{ij}^2}{r_{max}^2}(1 + 0.5e^{-\frac{\phi_{ij}^2}{\sigma_{\phi}}}w^{sa})
\end{equation}
\begin{equation}
w^{sa} \sim \mathcal{R}(\sigma^{sa})
\end{equation}
where $\phi_{ij}, r_{ij}$ are the azimuth and range of the $i, j$ bin. $r_{max}$ represents the maximum range of the sonar, $\sigma_{\phi}$ is used as a parameter to model the beam pattern-dependent noise gain, and $w^{sa}$ is sampled from a Rayleigh distribution with noise parameter $\sigma^{sa}$. We then compose the additive noise with multiplicative noise $w^{sm}$. 
\begin{equation}
\tilde{I}_{ij} = I_{ij}(0.5 + w^{sm}) + \tilde{w}^{sa} 
\end{equation}
\begin{equation}
w^{sm} \sim \mathcal{N}(0, \sigma^{sm})
\end{equation}

We set up the default parameters of our proposed sonar model based on a Blueprint Subsea Oculus M750-d sonar, which is a common multibeam imaging sonar used in prior works~\cite{loi2024sonar, spears2020ice, feng2023fish, ge2024advanced}. The Oculus M750-d used in our experiments has a 130$^\circ$ horizontal field-of-view and a 20$^\circ$ vertical field of view. To emulate its behavior to the best extent, we perform a range-wise normalization after collapsing the 3D intensity data onto the 2D grid. We provide user-friendly APIs to choose between various plot styles, binning, and normalization methods.

\subsubsection{DVL}
DVLs mainly serve to estimate the linear velocity of the robot for navigation. DVLs typically use a 4-beam convex Janus array as the transducer setup, which provides four sparse range measurements. As noted in~\cite{song2023uncertainty,song2024turtlmap}, a specific feature of DVL is its adaptive measurement rate, which introduces challenges for multi-sensor fusion. Additionally, when a certain number of beams are out of the operational range, sensor dropout can occur, where the DVL will return invalid velocity. We include these two sources of noise in our DVL model. This will enable users to test state estimation algorithms in the simulator with noisy but realistic DVL returns.

\begin{figure*}[t]
    \centering
    \includegraphics[width=1.0\linewidth]{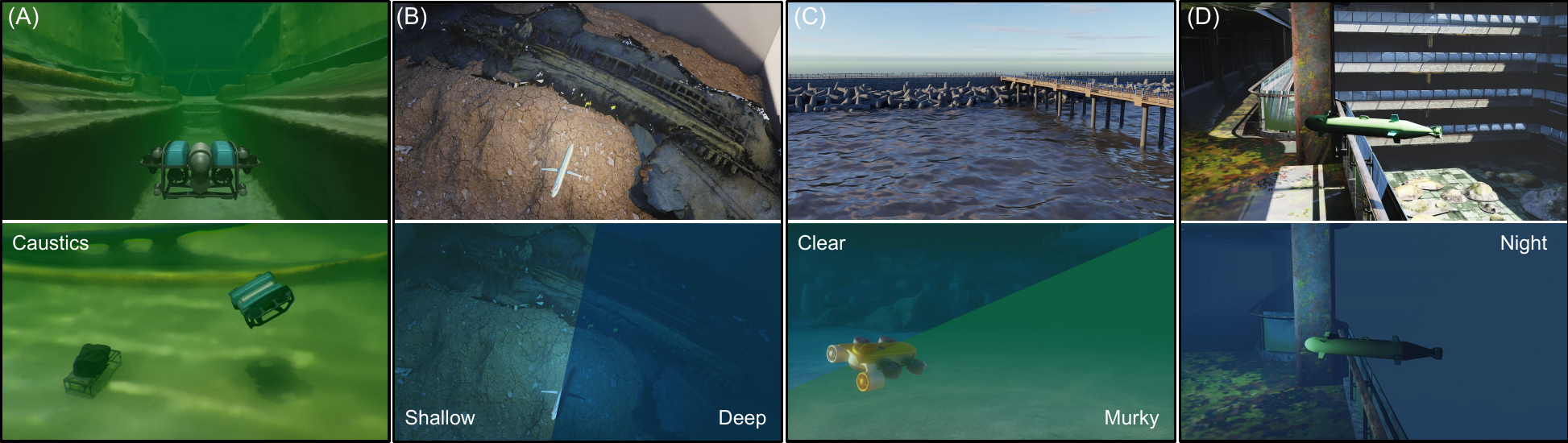}
    \caption{OceanSim offers flexibility to incorporate various scene and environment models. Details of constructing the MHL towing tank (A) and \textit{Monohansett} shipwreck (B) scenes are discussed in Sec.~\ref{sec:oceansim-mhl-env} and Sec.~\ref{sec:oceansim-shipwreck-env}. We also demonstrate OceanSim with online assets such as a pier (C)~\cite{pierModel} and an underwater mall (D)~\cite{mallModel} scene. In the second row, we show different water effects OceanSim can simulate for each of these sample scenes.}
    \label{fig:simulation_environments}
\end{figure*}

\subsection{GPU-accelerated Simulated Data Generation}
We design the rendering functions of the underwater camera and imaging sonar to fully leverage the highly efficient synthetic data generation capability of the NVIDIA Omniverse replicator tool. We also develop our rendering operations using the GPU-based parallel computing library (i.e., NVIDIA Warp) to enhance rendering speed. OceanSim allows the user to operate the robot, visualize sensor data, and record data simultaneously.

\subsection{Simulation Environments}
Being part of the NVIDIA Omniverse~\cite{Nvidia_omniverse} ecosystem that fully supports Universal Scene Description (OpenUSD)~\cite{OpenUSD}, the workflow for creating new simulation environments in OceanSim is highly efficient and customizable. Discussed in this sub-section, we test two environments for OceanSim that are generated with real-world data (Fig.~\ref{fig:simulation_environments}), including a water tank scene and a shipwreck scene. Note that the user can also import arbitrary large-scale environment models found online, as long as the generated file types are compatible with OpenUSD and NVIDIA Omniverse. To demonstrate this, we additionally test a simulated pier scene~\cite{pierModel} and an underwater mall scene~\cite{mallModel}. We will provide documentation on how to load additional environment models into OceanSim.%

\subsubsection{MHL Towing Tank}
\label{sec:oceansim-mhl-env}
We constructed a digital twin of the University of Michigan Marine Hydrodynamics Laboratory (MHL) towing tank~\cite{MHL_Tanks}. To construct the digital twin, when the tank was emptied, we took a survey scan with a stereo camera. We then used a photogrammetric tool~\cite{agisoft_metashape_2.2.0} to construct a 3D model with textures using the captured video log. The 3D model is imported into OceanSim as a simulation environment. As shown in Fig.~\ref{fig:simulation_environments}, this towing tank scene is low-texture and has substantial repetitive patterns, which will serve as a challenging playground to evaluate the robustness of underwater SLAM solutions. Notably, by creating a digital twin of a real test facility, we can collect real-world data and generate simulated data of the same scene to validate the accuracy and realism of OceanSim.%

\subsubsection{Monohansett Shipwreck}
\label{sec:oceansim-shipwreck-env}
 We provide another simulation environment example featuring a real-world shipwreck scene, the \textit{Monohansett} shipwreck site~\cite{monohansett}.  %
 The 3D model was computed from images collected by an underwater robot during prior field expeditions~\cite{sethuraman2024ai4shipwreck} using open-source 3D reconstruction software~\cite{alicevision2021}. %
 To design the shipwreck scene in Isaac Sim, we add a virtual container of the shipwreck and place the shipwreck model in a cuboid water tank. Within Isaac Sim, we further edit the bottom surface material to emulate sand. %
 The final shipwreck scene has a mixture of low-texture and high-texture regions, which is representative of many underwater sites of interest.

\subsection{Interface Design}
Since we design OceanSim as an extension package to NVIDIA Issac Sim~\cite{issacsim}, users can leverage other extension packages simultaneously. One natural benefit of being part of NVIDIA Issac Sim is that OceanSim directly supports ROS~\cite{ros} workflows. We also provide script tools as bridges for message types that are not supported by default. Additionally, we follow the development guidance of NVIDIA Omniverse and Isaac Sim so that users can use OceanSim in the same ways as other packages in NVIDIA Isaac Sim, further smoothing the learning curve. We provide detailed documentation and code examples in our public release.

\section{Experiment \& Results}
\begin{table*}[]
\centering
\caption{Quantitative comparison of RGB angular error. Lower is better. We sample two different views in which the camera has good visibility of the color board. We compare the rendered underwater image with HoloOcean~\cite{Holoocean}, UNav-Sim*~\cite{UNav_sim}, and the default camera model in NVIDIA Issac Sim~\cite{issacsim}. Note that UNav-Sim* uses Unreal Engine 5.1 to reproduce the image rendering pipeline from UNav-Sim~\cite{UNav_sim}. The unit is degree. \textbf{Bold} is best and \underline{underlined} is second-best.}
\label{tab:color board error}
\resizebox{1.0\textwidth}{!}{
\begin{tabular}{c|ccccccc|ccccccc}
\hline
\multirow{2}{*}{Method} & \multicolumn{7}{c|}{View 1} & \multicolumn{7}{c}{View 2} \\ \cline{2-15} 
                        & \cellcolor{yellow}\textcolor{black}{Yellow} 
                        & \cellcolor{cyan}\textcolor{white}{Cyan} 
                        & \cellcolor{green}\textcolor{black}{Green} 
                        & \cellcolor{magenta}\textcolor{black}{Magenta} 
                        & \cellcolor{red}\textcolor{black}{Red} 
                        & \cellcolor{blue}\textcolor{white}{Blue} 
                        & Mean 
                        & \cellcolor{yellow}\textcolor{black}{Yellow} 
                        & \cellcolor{cyan}\textcolor{white}{Cyan} 
                        & \cellcolor{green}\textcolor{black}{Green} 
                        & \cellcolor{magenta}\textcolor{black}{Magenta} 
                        & \cellcolor{red}\textcolor{black}{Red} 
                        & \cellcolor{blue}\textcolor{white}{Blue} 
                        & Mean \\ \hline
HoloOcean~\cite{Holoocean} & 14.76 & 15.10 & 20.27 & 19.95 & 17.42 & 26.55 & 19.01 & 15.59 & 10.63 & 16.62 & 16.91 & 16.34 & 24.23 & 16.72  \\
UNav-Sim*~\cite{UNav_sim} & \textbf{5.03} & \underline{11.06} & 9.88 & \underline{13.24} & \underline{7.47} & \underline{13.10} & \underline{9.96} & \textbf{10.76} & \underline{12.28} & 16.67 & \underline{13.22} & \textbf{6.48} & \textbf{16.24} & \textbf{12.61} \\
Isaac Sim                                        & 12.88  & 12.50 & \underline{3.85} & 19.77 & 18.51 & 22.32 & 14.97 & 18.64 & 14.93 & \textbf{15.29} & 25.04 & 28.24 & 32.20 & 22.39 \\
OceanSim (ours)                                         & \underline{8.32} & \textbf{7.00} & \textbf{3.80} & \textbf{9.72} & \textbf{7.06} & \textbf{11.39} & \textbf{7.88} & \underline{14.23} & \textbf{10.97} & \underline{15.86} & \textbf{10.09} & \underline{14.32} & \underline{17.18} & \underline{13.78} \\ \hline
\end{tabular}
}
\vspace{2mm}
\end{table*}

\begin{table}[]
\centering
\caption{Sonar rendering speed comparison. The cache entry represents the time cost for generating the octree cache for HoloOcean~\cite{Holoocean} in the unit of seconds. OceanSim does not require this process, offering better flexibility in simulation workflows. We select the MHL towing tank (Fig.~\ref{fig:simulation_environments}-A) as scene 1, the pier (Fig.~\ref{fig:simulation_environments}-C) as scene 2, and the underwater mall (Fig.~\ref{fig:simulation_environments}-D) as scene 3. $^*$~denotes that we use a coarser octree (0.04~m instead of default 0.02~m) to bypass a frozen error during octree generation.}

\label{tab:rendering-speed}
\begin{tabular}{c|cc|cc|cc}
\hline
\multirow{2}{*}{Method} & \multicolumn{2}{c|}{Scene 1} & \multicolumn{2}{c|}{Scene 2} & \multicolumn{2}{c}{Scene 3} \\ \cline{2-7} 
                        & Cache          & FPS         & Cache$^*$          & FPS         & Cache         & FPS         \\ \hline
HoloOcean~\cite{Holoocean}               & 417.8               & 3.8            & 317.0               & 3.3            &  502.1             & 1.5            \\
OceanSim (ours)                &  \textbf{0}              &  \textbf{31.7}           & \textbf{0}               & \textbf{30.3}            &  \textbf{0}             &    \textbf{28.2}        
\end{tabular}
\vspace{-2mm}
\end{table}

\subsection{Demonstration of High-fidelity Digital Twin}
\begin{figure}[b]
    \centering
    \includegraphics[width=1.0\linewidth]{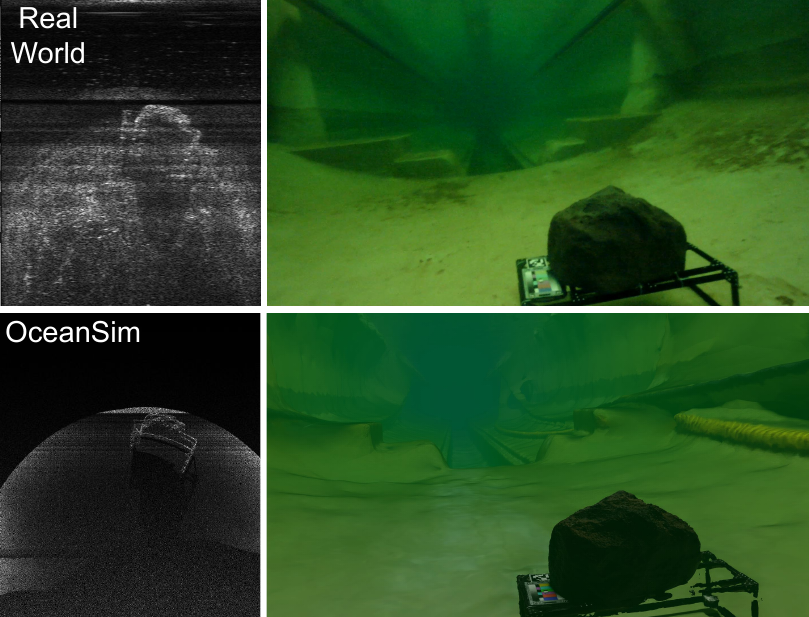}
    \caption{Qualitative comparison of real-world data (top) and the simulated sonar-image pair (bottom). For the above simulated sonar image, we use 3000 x 460 for the raycast density at horizontal and vertical FOVs, and 350 x 220 for the number of bins in range and azimuth directions.}
    \label{fig:qualitative_comparison_sonar_image}
    \vspace{-2mm}
\end{figure}

We conduct experiments to validate the advanced physics-based rendering techniques and sensor models proposed in OceanSim. Figure~\ref{fig:qualitative_comparison_sonar_image} presents qualitative results rendered by OceanSim compared with the sensor data captured in real-world experiments in the MHL towing tank. %

In our real experiment, we put a rock platform in the towing tank and operate the robot to capture the camera image and sonar scan. To replicate this experiment and obtain rendered sensor returns, we use the MHL 3D model discussed in Sec.~\ref{sec:oceansim-mhl-env} and place the 3D model of the rock platform at the same location as the real experiment. We initialize a camera (Sec.~\ref{sec:sensor-model-camera}) and a sonar (Sec.~\ref{sec:sensor-model-sonar}) in OceanSim using parameters that align with the actual sensors used. The qualitative comparison validates the effectiveness of OceanSim in constructing a high-fidelity digital twin, with rendered camera and sonar data that matches real-world data.

In addition, we also quantify the accuracy of our underwater image rendering model using RGB angular error ($\bar{\psi}$) as in~\cite{waternerf, hazelines}: 
\begin{equation} 
\bar{\psi}(c_1, c_2) = cos^{-1}\left[ \frac{(c_1 \cdot c_2)}{||c_1||||c_2||} \right]
\end{equation}

This metric compares two colors $c_1, c_2 \in \mathbb{R}^3$ and is invariant to lighting conditions via normalization. To compute this error, we utilize a colorboard in the scene. The colorboard consists of grayscale, yellow, cyan, green, magenta, red, and blue patches. We compare the reference underwater image from the real environment with the rendered images from OceanSim. We utilize a GUI to select multiple points on each patch, find the average pixel value, and compute $\bar{\psi}$ between the real reference image and the rendered image from OceanSim for each color patch. We report the average $\bar{\psi}$  across all color patches in Table~\ref{tab:color board error}. We provide comparison to underwater image rendering frameworks used in state-of-the-art underwater simulators including HoloOcean~\cite{Holoocean} and UNav-Sim~\cite{UNav_sim}, and the default camera model in NVIDIA Isaac Sim~\cite{issacsim}. We use the official release v1.0 of HoloOcean. Note that the underwater scene configuration presented in UNav-Sim is not publicly available, so we use the same version of Unreal Engine, UE5.1, to reproduce the underwater rendering effects leveraged in UNav-Sim, which we will refer to as UNav-Sim*. For all environments, we manually tune the underwater imaging parameters to qualitatively match the appearance of the real underwater images.
Results demonstrate that our proposed model consistently achieves best or second best in RGB angular error across all entries, validating the accuracy of our underwater image rendering pipeline. %

\subsection{Quantitative Comparison of Sonar Rendering Efficiency}
One of the core contributions of OceanSim is the accelerated rendering speed for sonar images, which is essential for generating large amounts of simulated sonar data to support training for data-driven perception models. To validate this feature, we quantitatively compare the sonar rendering speed of OceanSim with HoloOcean~\cite{Holoocean}. %

To ensure a fair comparison of time cost, all experiments on both HoloOcean and OceanSim are conducted on the same desktop, which has an NVIDIA RTX A6000 GPU, AMD Ryzen 5950X CPU, and $128$G RAM. We also unify the scenes and sensor configurations in both simulators. It is worth mentioning that we are not able to export the default simulation environments used in HoloOcean~\cite{Holoocean}. Therefore, we follow the documentation to import the scenes selected for comparison into HoloOcean using the Unreal Engine editor. We select three example scenes, including the MHL towing tank, a pier, and a simulated underwater mall scene. In practice, to quantify the rendering speed, we place the sensor rig at the same location in both simulators and tick the simulator to generate $100$ frames of sensor measurements. As discussed in Sec.~\ref{sec:related-works-underwater-simulator}, HoloOcean~\cite{Holoocean} requires building a partial octree on first startup of a new scene. We also report the time needed to build the cache of the partial octree. In the follow-up sensor data generation, since the sensor rig is kept static, we ensure the quantified rendering speed of HoloOcean directly represents its best performance with pre-built cache.

Table~\ref{tab:rendering-speed} demonstrates that OceanSim significantly outperforms HoloOcean in sonar rendering speed. Notably, OceanSim benefits from GPU-based ray tracing to achieve real-time sonar rendering speed without requiring extra time for building cache. This feature is important to enabling large-scale simulated data generation for underwater robotics applications. We also observe that the rendering speed is more impacted by scaling up the resolution (i.e. the number of bins) versus increasing the raycast density, and overly increasing resolutions can lead to fidelity drop due to the rendered images appearing much sharper than the real world ones. %

\section{Limitations \& Future Work}

Despite the notable improvement on physics-based underwater sensor modeling, rendering quality, efficiency, and flexibility of 3D workflows, OceanSim still has areas for improvement that we will aim to address in future development. First, the current version of OceanSim is a perception-oriented simulator, which means it does not accurately model underwater vehicle dynamics and fluid dynamics. Though we have observed empirical success of adding external force to the robot to simulate Fossen dynamics~\cite{Fossen} in OceanSim following~\cite{Holoocean}, this method is sub-optimal and does not yet fully leverage the physics simulation capability of NVIDIA Isaac Sim. We also do not model the water fluid dynamics and their effect on the robot, which will be critical towards building a full-stack underwater robotics simulator. We hope to address these limitations in the future to enable direct sim-to-real transfer of a controls policy model trained using OceanSim. In addition, the current beta release of OceanSim assumes a user group of mostly developers, so configuration of new simulation scenarios is done through coding. Developing a graphical user interface of OceanSim sensors  in NVIDIA Isaac Sim will greatly improve the accessibility of OceanSim and underwater robotics research. Lastly, OceanSim does not implement optical and acoustic communication modems to simulate communications between cooperative agents. We value the importance of this feature and think extending OceanSim to support it will be an interesting future direction. As an open-source project, we are committed to addressing these challenges in future updates and invite the community to join us in refining and advancing OceanSim for the benefit of all underwater robotics research.

\section{Conclusion}
We have proposed OceanSim, a high-performance underwater simulation framework. We demonstrate the effectiveness of rendering techniques and sensor models proposed by OceanSim through qualitative and quantitative comparison with real-world data. We additionally test the accelerated sonar rendering speed of OceanSim via extensive evaluation against the state-of-the-art underwater simulator, validating the enhanced scalability of OceanSim. We have also demonstrated the highly flexible 3D workflows that OceanSim supports, and will release OceanSim with detailed documentation to support the marine robotics research community.

\section{Acknowledgement}
We would like to acknowledge the contributions made by previous underwater simulators as they are the foundations of this work. We also thank the contributors of NVIDIA Issac Sim and Omniverse and OpenUSD. We appreciate the support for real-world experiments from all of the staff at the University of Michigan Marine Hydrodynamics Laboratory. We also thank the iCORE Lab at Louisiana State University led by Prof. Corina Barbalata for their work on reconstructing the \textit{Monohansett} shipwreck.
A GenAI tool (i.e., GPT-4o) was used to prepare the initial draft of Sec.~\ref{Sec:introduction}.

\bibliographystyle{IEEEtran}
\bibliography{reference}

\begin{thebibliography}{10}
\providecommand{\url}[1]{#1}
\csname url@samestyle\endcsname
\providecommand{\newblock}{\relax}
\providecommand{\bibinfo}[2]{#2}
\providecommand{\BIBentrySTDinterwordspacing}{\spaceskip=0pt\relax}
\providecommand{\BIBentryALTinterwordstretchfactor}{4}
\providecommand{\BIBentryALTinterwordspacing}{\spaceskip=\fontdimen2\font plus
\BIBentryALTinterwordstretchfactor\fontdimen3\font minus \fontdimen4\font\relax}
\providecommand{\BIBforeignlanguage}[2]{{%
\expandafter\ifx\csname l@#1\endcsname\relax
\typeout{** WARNING: IEEEtran.bst: No hyphenation pattern has been}%
\typeout{** loaded for the language `#1'. Using the pattern for}%
\typeout{** the default language instead.}%
\else
\language=\csname l@#1\endcsname
\fi
#2}}
\providecommand{\BIBdecl}{\relax}
\BIBdecl

\bibitem{rahman2019svin2}
S.~Rahman, A.~Q. Li, and I.~Rekleitis, ``Svin2: An underwater slam system using sonar, visual, inertial, and depth sensor,'' in \emph{2019 IEEE/RSJ International Conference on Intelligent Robots and Systems (IROS)}.\hskip 1em plus 0.5em minus 0.4em\relax IEEE, 2019, pp. 1861--1868.

\bibitem{zhang2024recgs}
T.~Zhang, W.~Zhi, B.~Meyers, N.~Durrant, K.~Huang, J.~Mangelson, C.~Barbalata, and M.~Johnson-Roberson, ``Recgs: Removing water caustic with recurrent gaussian splatting,'' \emph{IEEE Robotics and Automation Letters}, 2024.

\bibitem{zhao2023tightly}
L.~Zhao, M.~Zhou, and B.~Loose, ``Tightly-coupled visual-dvl-inertial odometry for robot-based ice-water boundary exploration,'' in \emph{2023 IEEE/RSJ International Conference on Intelligent Robots and Systems (IROS)}.\hskip 1em plus 0.5em minus 0.4em\relax IEEE, 2023, pp. 7127--7134.

\bibitem{kim2013real}
A.~Kim and R.~M. Eustice, ``Real-time visual slam for autonomous underwater hull inspection using visual saliency,'' \emph{IEEE Transactions on Robotics}, vol.~29, no.~3, pp. 719--733, 2013.

\bibitem{bagoren2025pugs}
O.~Bagoren, M.~Micatka, K.~A. Skinner, and A.~Marburg, ``Pugs: Perceptual uncertainty for grasp selection in underwater environments,'' in \emph{2025 IEEE International Conference on Robotics and Automation (ICRA)}.\hskip 1em plus 0.5em minus 0.4em\relax IEEE, 2025.

\bibitem{Holoocean}
E.~Potokar, K.~Lay, K.~Norman, D.~Benham, S.~Ashford, R.~Peirce, T.~B. Neilsen, M.~Kaess, and J.~G. Mangelson, ``Holoocean: A full-featured marine robotics simulator for perception and autonomy,'' \emph{IEEE Journal of Oceanic Engineering}, vol.~49, no.~4, pp. 1322--1336, 2024.

\bibitem{potokar2022holoocean}
E.~Potokar, K.~Lay, K.~Norman, D.~Benham, T.~B. Neilsen, M.~Kaess, and J.~G. Mangelson, ``Holoocean: Realistic sonar simulation,'' in \emph{2022 IEEE/RSJ International Conference on Intelligent Robots and Systems (IROS)}.\hskip 1em plus 0.5em minus 0.4em\relax IEEE, 2022, pp. 8450--8456.

\bibitem{UNav_sim}
A.~Amer, O.~{\'A}lvarez-Tu{\~n}{\'o}n, H.~{\.I}. U{\u{g}}urlu, J.~L.~F. Sejersen, Y.~Brodskiy, and E.~Kayacan, ``Unav-sim: A visually realistic underwater robotics simulator and synthetic data-generation framework,'' in \emph{2023 21st International Conference on Advanced Robotics (ICAR)}, 2023, pp. 570--576.

\bibitem{UWsim}
M.~Prats, J.~Pérez, J.~J. Fernández, and P.~J. Sanz, ``An open source tool for simulation and supervision of underwater intervention missions,'' in \emph{2012 IEEE/RSJ International Conference on Intelligent Robots and Systems}, 2012, pp. 2577--2582.

\bibitem{ros}
M.~Quigley, B.~Gerkey, K.~Conley, J.~Faust, T.~Foote, J.~Leibs, E.~Berger, R.~Wheeler, and A.~Ng, ``Ros: an open-source robot operating system,'' in \emph{Proc. of the IEEE Intl. Conf. on Robotics and Automation (ICRA) Workshop on Open Source Robotics}, 2009.

\bibitem{UUV_sim}
M.~M.~M. Manhães, S.~A. Scherer, M.~Voss, L.~R. Douat, and T.~Rauschenbach, ``Uuv simulator: A gazebo-based package for underwater intervention and multi-robot simulation,'' in \emph{OCEANS 2016 MTS/IEEE Monterey}, 2016, pp. 1--8.

\bibitem{UW_Morse}
E.~H. Henriksen, I.~Schjølberg, and T.~B. Gjersvik, ``Uw morse: The underwater modular open robot simulation engine,'' in \emph{2016 IEEE/OES Autonomous Underwater Vehicles (AUV)}, 2016, pp. 261--267.

\bibitem{URsim}
M.~Sewtz, H.~Lehner, Y.~Fanger, J.~Eberle, M.~Wudenka, M.~G. Müller, T.~Bodenmüller, and M.~J. Schuster, ``Ursim - a versatile robot simulator for extra-terrestrial exploration,'' in \emph{2022 IEEE Aerospace Conference (AERO)}, 2022, pp. 1--14.

\bibitem{Stonefish}
P.~Cieślak, ``Stonefish: An advanced open-source simulation tool designed for marine robotics, with a ros interface,'' in \emph{OCEANS 2019 - Marseille}, 2019, pp. 1--6.

\bibitem{sim2vita}
P.~D. de~Cerqueira~Gava, C.~L. Nascimento~Júnior, J.~R. Belchior~de França~Silva, and G.~J. Adabo, ``Simu2vita: A general purpose underwater vehicle simulator,'' \emph{Sensors}, vol.~22, no.~9, p. 3255, 2022.

\bibitem{unrealengine}
{Epic Games}, ``Unreal engine,'' \url{https://www.unrealengine.com}.

\bibitem{issacsim}
J.~Liang, V.~Makoviychuk, A.~Handa, N.~Chentanez, M.~Macklin, and D.~Fox, ``Gpu-accelerated robotic simulation for distributed reinforcement learning,'' in \emph{Conference on Robot Learning}.\hskip 1em plus 0.5em minus 0.4em\relax PMLR, 2018, pp. 270--282.

\bibitem{Nvidia_omniverse}
``Nvidia omniverse,'' \url{https://docs.nvidia.com/omniverse/index.html}, accessed: 2024-12-01.

\bibitem{gaussianSplatting_Isaac}
X.~Li, J.~Li, Z.~Zhang, R.~Zhang, F.~Jia, T.~Wang, H.~Fan, K.-K. Tseng, and R.~Wang, ``Robogsim: A real2sim2real robotic gaussian splatting simulator,'' \emph{arXiv preprint arXiv:2411.11839}, 2024.

\bibitem{object_detection_Isaac}
X.~Chen, Y.~Li, and H.~Lu, ``Few-shot object detection algorithm based on geometric prior and attention rpn,'' in \emph{2024 International Wireless Communications and Mobile Computing (IWCMC)}, 2024, pp. 706--711.

\bibitem{DRL_Isaac}
M.~Rojas, G.~Hermosilla, D.~Yunge, and G.~Farias, ``An easy to use deep reinforcement learning library for ai mobile robots in isaac sim,'' \emph{Applied Sciences}, vol.~12, no.~17, p. 8429, 2022.

\bibitem{RL_Isaac}
E.~Berrocal, B.~Sierra, and H.~Herrero, ``Evaluating pybullet and isaac sim in the scope of robotics and reinforcement learning,'' in \emph{2024 7th Iberian Robotics Conference (ROBOT)}, 2024, pp. 1--6.

\bibitem{humanoid_RL_Isaac}
X.~Gu, Y.-J. Wang, and J.~Chen, ``Humanoid-gym: Reinforcement learning for humanoid robot with zero-shot sim2real transfer,'' 2024.

\bibitem{PBS_simulation}
E.~Angelidis, J.~Arreguit, J.~Bender, P.~Berggold, Z.~Liu, A.~Knoll, A.~Crespi, and A.~J. Ijspeert, ``A smoothed particle hydrodynamics framework for fluid simulation in robotics,'' \emph{Robotics and Autonomous Systems}, vol. 185, p. 104885, 2025.

\bibitem{fluid_Gazebo}
L.~Chen, R.~Cui, W.~Yan, C.~Yang, Z.~Li, H.~Xu, and H.~Yu, ``Stability criterion and stability enhancement for a thruster-assisted underwater hexapod robot,'' \emph{IEEE Transactions on Robotics}, vol.~41, pp. 42--61, 2025.

\bibitem{Automation_production_Isaac}
S.~Nambiar, M.~Jonsson, and M.~Tarkian, ``Automation in unstructured production environments using isaac sim: A flexible framework for dynamic robot adaptability,'' \emph{Procedia CIRP}, vol. 130, pp. 837--846, 2024.

\bibitem{automation_factory_Isaac}
A.~Haroon, A.~Lakshman, M.~Mundy, and B.~Li, ``Autonomous robotic 3d scanning for smart factory planning,'' in \emph{Dimensional Optical Metrology and Inspection for Practical Applications XIII}, vol. 13038.\hskip 1em plus 0.5em minus 0.4em\relax SPIE, 2024, pp. 104--112.

\bibitem{manipulation_Isaac}
D.~H. Nguyen, T.~Schneider, G.~Duret, A.~Kshirsagar, B.~Belousov, and J.~Peters, ``Tacex: Gelsight tactile simulation in isaac sim--combining soft-body and visuotactile simulators,'' \emph{arXiv preprint arXiv:2411.04776}, 2024.

\bibitem{manipulation_training_Isaac}
Y.~Son, H.~Han, and J.~Cho, ``Usefulness of using nvidia isaacsim and isaacgym for ai robot manipulation training,'' in \emph{2023 14th International Conference on Information and Communication Technology Convergence (ICTC)}, 2023, pp. 1725--1728.

\bibitem{grasp_Isaac}
L.~F. Casas, N.~Khargonkar, B.~Prabhakaran, and Y.~Xiang, ``Multigrippergrasp: A dataset for robotic grasping from parallel jaw grippers to dexterous hands,'' in \emph{2024 IEEE/RSJ International Conference on Intelligent Robots and Systems (IROS)}.\hskip 1em plus 0.5em minus 0.4em\relax IEEE, 2024.

\bibitem{semantics_Isaac}
H.~X. Zhang and Z.~Zou, ``Quality assurance for building components through point cloud segmentation leveraging synthetic data,'' \emph{Automation in Construction}, vol. 155, p. 105045, 2023.

\bibitem{zhang2025xpg}
Y.~Zhang, S.~Li, and E.~Shrestha, ``Xpg-rl: Reinforcement learning with explainable priority guidance for efficiency-boosted mechanical search,'' 2025.

\bibitem{OpenUSD}
P.~A. Studios \emph{et~al.}, ``Universal scene description (openusd),'' \url{https://openusd.org/}.

\bibitem{chu2024marinegym}
S.~Chu, Z.~Huang, M.~Lin, D.~Li, and I.~Carlucho, ``Marinegym: Accelerated training for underwater vehicles with high-fidelity rl simulation,'' 2024.

\bibitem{akkaynak_what_2017}
D.~Akkaynak, T.~Treibitz, T.~Shlesinger, Y.~Loya, R.~Tamir, and D.~Iluz, ``What is the space of attenuation coefficients in underwater computer vision?'' in \emph{2017 IEEE Conference on Computer Vision and Pattern Recognition (CVPR)}, 2017, pp. 568--577.

\bibitem{akkaynak_revised_2018}
D.~Akkaynak and T.~Treibitz, ``A revised underwater image formation model,'' in \emph{2018 IEEE/CVF Conference on Computer Vision and Pattern Recognition}, 2018, pp. 6723--6732.

\bibitem{OmniverseReplicator}
NVIDIA, ``Omniverse replicator,'' \url{https://docs.omniverse.nvidia.com/extensions/latest/ext_replicator.html}.

\bibitem{raytrace_sonar_sim}
R.~Cerqueira, T.~Trocoli, G.~Neves, S.~Joyeux, J.~Albiez, and L.~Oliveira, ``A novel gpu-based sonar simulator for real-time applications,'' \emph{Computers \& Graphics}, vol.~68, pp. 66--76, 2017.

\bibitem{demarco2015computationally}
K.~J. DeMarco, M.~E. West, and A.~M. Howard, ``A computationally-efficient 2d imaging sonar model for underwater robotics simulations in gazebo,'' in \emph{OCEANS 2015-MTS/IEEE Washington}.\hskip 1em plus 0.5em minus 0.4em\relax IEEE, 2015, pp. 1--7.

\bibitem{loi2024sonar}
N.~Loi, Y.~Z. Tan, E.~W. Goh, and M.~H. Ang, ``Sonar slam in structured underwater environments,'' in \emph{OCEANS 2024-Singapore}.\hskip 1em plus 0.5em minus 0.4em\relax IEEE, 2024, pp. 1--10.

\bibitem{spears2020ice}
A.~Spears, M.~Meister, C.~Ramey, J.~Lawrence, D.~Dichek, P.~Washam, and B.~E. Schmidt, ``Ice topography reconstruction and panoramic stitching using forward looking sonar images,'' in \emph{Global Oceans 2020: Singapore--US Gulf Coast}.\hskip 1em plus 0.5em minus 0.4em\relax IEEE, 2020, pp. 1--6.

\bibitem{feng2023fish}
Y.~Feng, Y.~Wei, S.~Sun, J.~Liu, D.~An, and J.~Wang, ``Fish abundance estimation from multi-beam sonar by improved mcnn,'' \emph{Aquatic Ecology}, vol.~57, no.~4, pp. 895--911, 2023.

\bibitem{ge2024advanced}
L.~Ge, P.~Singh, and A.~Sadhu, ``Advanced deep learning framework for underwater object detection with multibeam forward-looking sonar,'' \emph{Structural Health Monitoring}, p. 14759217241235637, 2024.

\bibitem{song2023uncertainty}
J.~Song, O.~Bagoren, and K.~A. Skinner, ``Uncertainty-aware acoustic localization and mapping for underwater robots,'' in \emph{OCEANS 2023-Limerick}.\hskip 1em plus 0.5em minus 0.4em\relax IEEE, 2023, pp. 1--9.

\bibitem{song2024turtlmap}
J.~Song, O.~Bagoren, R.~Andigani, A.~Sethuraman, and K.~A. Skinner, ``Turtlmap: Real-time localization and dense mapping of low-texture underwater environments with a low-cost unmanned underwater vehicle,'' in \emph{2024 IEEE/RSJ International Conference on Intelligent Robots and Systems (IROS)}, 2024, pp. 1191--1198.

\bibitem{pierModel}
``Hydro pier-bridge,'' \url{https://skfb.ly/oroqH}, sketchfab asset.

\bibitem{mallModel}
``Ruins of the underwater shopping mall,'' \url{https://skfb.ly/oGSVJ}, sketchfab asset.

\bibitem{MHL_Tanks}
``Aaron friedman marine hydrodynamics laboratory,'' \url{https://mhl.engin.umich.edu}.

\bibitem{agisoft_metashape_2.2.0}
{Agisoft LLC}, \emph{Agisoft Metashape 2.2.0 Professional Edition}, 2024, photogrammetric processing software. Available at \url{https://www.agisoft.com/}.

\bibitem{monohansett}
``Monohansett,'' \url{https://thunderbay.noaa.gov/shipwrecks/monohansett.html}, monohansett shipwreck.

\bibitem{sethuraman2024ai4shipwreck}
A.~V. Sethuraman, A.~Sheppard, O.~Bagoren, C.~Pinnow, J.~Anderson, T.~C. Havens, and K.~A. Skinner, ``Machine learning for shipwreck segmentation from side scan sonar imagery: Dataset and benchmark,'' \emph{The International Journal of Robotics Research}, p. 02783649241266853, 2024.

\bibitem{alicevision2021}
C.~Griwodz, S.~Gasparini, L.~Calvet, P.~Gurdjos, F.~Castan, B.~Maujean, G.~D. Lillo, and Y.~Lanthony, ``Alicevision meshroom: An open-source 3d reconstruction pipeline,'' in \emph{Proceedings of the 12th ACM Multimedia Systems Conference - MMSys '21}.\hskip 1em plus 0.5em minus 0.4em\relax ACM Press, 2021.

\bibitem{waternerf}
A.~V. Sethuraman, M.~S. Ramanagopal, and K.~A. Skinner, ``Waternerf: Neural radiance fields for underwater scenes,'' in \emph{OCEANS 2023 - MTS/IEEE U.S. Gulf Coast}, 2023, pp. 1--7.

\bibitem{hazelines}
D.~Akkaynak and T.~Treibitz, ``Sea-thru: A method for removing water from underwater images,'' in \emph{2019 IEEE/CVF Conference on Computer Vision and Pattern Recognition (CVPR)}, 2019, pp. 1682--1691.

\bibitem{Fossen}
``Fossen dynamics,'' \url{https://github.com/cybergalactic/PythonVehicleSimulator}, accessed: 2024-12-01.

\end{thebibliography}

\end{document}